\documentclass{article} 
\usepackage[preprint]{automl}

\usepackage{amsmath,amsfonts,bm}

\def\1{\bm{1}}

\DeclareMathAlphabet{\mathsfit}{\encodingdefault}{\sfdefault}{m}{sl}
\SetMathAlphabet{\mathsfit}{bold}{\encodingdefault}{\sfdefault}{bx}{n}

\PassOptionsToPackage{sort,square,numbers}{natbib}
\usepackage{amsfonts,mathtools}       
\usepackage{nicefrac}        
\usepackage{microtype}      
\usepackage{wrapfig}
\usepackage{svg}
\usepackage{xspace}
\usepackage{dsfont}
\usepackage{enumitem} 
\usepackage{color, colortbl}
\usepackage{pifont}
\usepackage{graphicx}
\usepackage{epstopdf}
\usepackage{subcaption}
\usepackage{pgfplots}
\usepackage{pgfplotstable}
\usepackage{comment}
\usepackage[capitalise]{cleveref}
\usepackage{siunitx}
\usepackage{listings}
\usepackage[utf8]{inputenc}
\usepackage{tabu}
\usepackage{array}
\usepackage{multirow}
\usepackage{diagbox}
\usepackage{placeins}
\usepackage{algorithm,lipsum,changepage}
\usepackage{float}
\usepackage{balance}
\usepackage{calc}
\usepackage{ifthen}
\usepackage{hyphenat}
\usepackage{acronym}
\pgfplotsset{compat=1.18}

\renewcommand{\epsilon}{\varepsilon}
\usepackage{algorithmic}

\newcommand{\new}[1]{#1}

\acrodef{BO}[BO]{Bayesian optimization}
\acrodef{GP}[GP]{Gaussian process}
\acrodef{MI}[MI]{mutual information}
\acrodef{GMM}[GMM]{Gaussian mixture model}
\acrodef{GT}[GT]{group testing}
\acrodef{SMC}[SMC]{sequential Monte Carlo}
\acrodef{RV}[RV]{random variable}

\newcommand{\baxus}{\texttt{BAxUS}\xspace}
\newcommand{\bounce}{\texttt{Bounce}\xspace}
\newcommand{\hesbo}{\texttt{HeSBO}\xspace}
\newcommand{\turbo}{\texttt{TuRBO}\xspace}
\newcommand{\alebo}{\texttt{Alebo}\xspace}
\newcommand{\rembo}{\texttt{REMBO}\xspace}

\newcommand{\saasbo}{\texttt{SAASBO}\xspace}
\newcommand{\cmaes}{\texttt{CMA-ES}\xspace}
\newcommand{\mctvs}{\texttt{MCTS-VS}\xspace}
\newcommand{\gtbo}{\texttt{GTBO}\xspace}
\newcommand{\xgboost}{\texttt{XGBoost}\xspace}
\newcommand{\fanova}{\texttt{fAnova}\xspace}
\newcommand{\vsbo}{\texttt{VS-BO}\xspace}

\title{Leveraging Axis-Aligned Subspaces for High-Dimensional Bayesian Optimization with Group Testing}

\author[1,2]{\nameemail{Erik Hellsten*}{erik.hellsten@cs.lth.se}}
\author[1]{\nameemail{Carl Hvarfner*}{carl.hvarfner@cs.lth.se}}
\author[1]{\nameemail{Leonard Papenmeier*}{leonard.papenmeier@cs.lth.se}}
\author[1,2]{\nameemail{Luigi Nardi}{luigi.nardi@cs.lth.se}}

\affil[1]{Lund University}
\affil[2]{DBtune}

\hypersetup{%
  pdfauthor={}, 
  pdftitle={},
  pdfsubject={},
  pdfkeywords={}
}

\begin{document}

\maketitle

\begin{abstract}
\ac{BO} is an effective method for optimizing expensive-to-evaluate black-box functions. While high-dimensional problems can be particularly challenging, due to the multitude of parameter choices and the potentially high number of data points required to fit the model, this limitation can be addressed if the problem satisfies simplifying assumptions. Axis-aligned subspace approaches, where few dimensions have a significant impact on the objective, motivated several algorithms for high-dimensional \ac{BO}. However, the validity of this assumption is rarely verified, and the assumption is rarely exploited to its full extent.
We propose a \ac{GT} approach to identify active variables to facilitate efficient optimization in these domains.
The proposed algorithm, Group Testing Bayesian Optimization (\gtbo), first runs a testing phase where groups of variables are systematically selected and tested on whether they influence the objective, then terminates once active dimensions are identified.
To that end, we extend the well-established \ac{GT} theory to functions over continuous domains.
In the second phase, \gtbo guides optimization by placing more importance on the active dimensions.
By leveraging the axis-aligned subspace assumption, \gtbo outperforms state-of-the-art methods on benchmarks satisfying the assumption of axis-aligned subspaces, while offering improved interpretability.
\end{abstract}

\section{Introduction}
Noisy and expensive-to-evaluate black-box functions occur in many practical optimization tasks, including material and chemical design~\cite{zhang2020bayesian, yik2024towards}, hardware design~\cite{nardi2019practical,ejjeh2022hpvm2fpga}, hyperparameter tuning~\cite{kandasamy2018neural, ru2020interpretable,chen_arXiv18, NEURIPS2023_fa55bf19}, and robotics~\cite{calandra-lion14a, berkenkamp2021safety, mayr2022learning}.
\ac{BO} is an established framework that allows optimization of such problems in a sample-efficient manner~\cite{shahriari-aistats16a,frazier2018tutorial}. Particularly challenging are high-dimensional domains such as robotics~\cite{calandra-lion14a}, drug discovery~\cite{negoescu2011knowledge} and vehicle design~\cite{jones2008large}.

In recent years, efficient approaches have been proposed to tackle the limitations of \ac{BO} in high dimensions. While the standard BO algorithm is highly efficient with reasonable assumptions~\cite{pmlr-v235-hvarfner24a}, many approaches assume the existence of a low-dimensional \textit{active subspace} of the input domain that has a significantly larger impact on the optimization objective than its complement~\cite{wang2016bayesian, letham2020re}. Furthermore, it is often assumed that the active subspace is \emph{axis-aligned}~\cite{nayebi2019framework, eriksson2021high, song2022monte, papenmeier2022increasing, papenmeier2023bounce} - that is, only a subset of the input variables (aligned with coordinate axes) significantly influence the objective function. This assumption simplifies modeling and can be effective in practice, but may now hold in real-world applications. To mitigate potential failures due to misaligned subspaces, several recent approaches have proposed relaxing the axis-alignment constraint~\cite{eriksson2021high, papenmeier2022increasing, papenmeier2023bounce}.

However, such relaxations come with trade-offs: they often reduce sample efficiency when the axis-aligned assumption does hold, and they may obscure interpretability by making it harder to identify which variables truly matter. Thus, we propose a complementary approach that \emph{aggressively exploits axis-alignment}. Specifically, we introduce a Group Testing Bayesian Optimization (GTBO) framework that serves as a true complement to the vanilla algorithm~\cite{pmlr-v235-hvarfner24a}, made for settings where the axis-aligned active subspace assumption is valid. When the active subspace is axis-aligned, finding the active dimensions can be framed as a feature selection problem.
A straightforward approach is first to learn the active dimensions using a dedicated feature selection approach and subsequently optimize over the learned subspace. We propose to initially find the active dimensions using an information-theoretic approach built around the well-established theory of \ac{GT}~\cite{dorfman1943detection}. \ac{GT} is the problem of finding several active elements within a larger set by iteratively testing groups of elements. 
We develop the theory needed to transition noisy \ac{GT}, which otherwise only allows binary observations, to support evaluations of continuous black-box functions drawn from a GP prior. This enables \ac{GT} in \ac{BO} and other applications, such as feature selection for regression problems.
The contributions of this work are:
\begin{enumerate}[leftmargin=*,noitemsep]
    \item We extend the theory of \acl{GT} to feature importance analysis in a continuous setting tailored towards \ac{GP} modeling.
    \item We introduce Group Testing Bayesian Optimization (\gtbo), a unified framework that integrates active learning and \ac{BO}. Assuming an axis-aligned active subspace, \gtbo uses activeness information extracted during an initial \ac{GT} phase to inform the subsequent optimization process.
    \item We demonstrate that \gtbo outperforms axis-aligned subspace BO methods and reliably identifies active dimensions with high probability.
\end{enumerate}

\section{Background}

\subsection{High-dimensional Bayesian optimization}
We aim to find a minimizer $\bm{x}^* \in \arg\min_{\bm{x} \in \mathcal{X}} f(\bm{x})$ of a black-box function $f : \mathcal{X} \rightarrow \mathbb{R}$, defined over a $D$-dimensional domain $\mathcal{X} = [0, 1]^D$. The function $f$ can only be queried point-wise, and observations are noisy: $y(\bm{x}) = f(\bm{x}) + \varepsilon$, where $\varepsilon \sim \mathcal{N}(0, \sigma_n^2)$ and $\sigma_n^2$ denotes the \emph{noise variance}. Lastly, we assume that $f$ is costly to evaluate, so the number of evaluations is limited.

Bayesian Optimization (BO) is a sample-efficient framework well-suited for such settings. BO typically models the unknown function $f$ using a Gaussian Process (GP) prior~\cite{10.5555/1162254}, which provides a posterior distribution over functions conditioned on observed data. GPs are particularly attractive due to their ability to quantify predictive uncertainty and their closed-form posterior updates. However, the performance of GPs deteriorates in high-dimensional spaces, leading to inefficiencies in both modeling and optimization. This phenomenon is commonly referred to as the \emph{curse of dimensionality} (CoD), which denotes the exponential increase in volume and sparsity of data as dimensionality grows. In high-dimensional Bayesian optimization (HDBO), CoD poses significant challenges for both GP inference and the acquisition function optimization. 
In this work, we consider high-dimensional optimization problems where only a small subset of the $D$ input dimensions, denoted by $d_e \ll D$, are \emph{effectively active}, while the remaining ${D - d_e}$ dimensions are \emph{inactive}. That is, the function $f$ can be well-approximated by varying only the active dimensions. This structure enables dimensionality-aware approaches that avoid global optimization over the full input space. Recent work~\cite{pmlr-v235-hvarfner24a} has demonstrated that efficient high-dimensional GP modeling and BO can be achieved by adjusting the lengthscale prior of the GP, thereby alleviating the CoD.

\subsection{Low-dimensional subspace Bayesian optimization}
Using linear embeddings is a common approach when optimizing high-dimensional functions that contain a low-dimensional active subspace. 
\rembo~\cite{wang2016bayesian} shows that a random embedded subspace with at least the same dimensionality as the active subspace is guaranteed to contain an optimum if the subspace is unbounded.
However, \ac{BO} usually requires a bounded search space, and \rembo suffers from projecting outside of this search space.
\hesbo~\cite{nayebi2019framework} uses a sparse projection matrix to avoid points outside the search space.
\baxus~\cite{papenmeier2022increasing} and \bounce~\cite{papenmeier2023bounce} use an \hesbo-like embedding~\cite{nayebi2019framework} but allow the dimensionality  to grow over time.
This ensures that the optimum can eventually be found but leads to \baxus and \bounce optimizing over high-dimensional spaces in later optimization stages.
\alebo~\cite{letham-ba18a} presents another remedy to shortcomings in the search space design of \rembo.
In particular, bounds from the original space are projected into the embedded space, and the kernel in the embedded space is adjusted to preserve distances from the original space.
 
\paragraph{Axis-aligned active subspaces}
%
One common assumption to tackle high-dimensional problems is that the active subspace is axis-aligned, i.e., a subspace that can be obtained by removing the inactive dimensions.
This is equivalent to the assumption of active and inactive dimensions.
\saasbo~\cite{eriksson2021high} sets up on this assumption by adding a strong sparsity-inducing prior to the hyperparameters of the \ac{GP} model, prioritizing fewer active dimensions unless the data strongly suggests otherwise.
%
\vsbo~\cite{shen2023computationally} actively identifies relevant variables in a problem, similar to our approach. However, it relies on a heuristic tied to a specific surrogate model (\ac{GP}). 
It lacks a clear-cut decision on variable relevance due to ongoing variable importance estimation during optimization.
\mctvs~\cite{song2022monte} uses a Monte Carlo tree search to select the active dimensions dynamically.
It relies on randomly chosen sets of active and inactive dimensions, making finding the ``correct'' active dimensions difficult if the number of active dimensions is high.
Our work also leverages the axis-aligned assumption, but it differs in how we identify the active dimensions. 

\paragraph{Active subspace learning}
In this paper, we resolve to a more direct approach, where we learn the active subspace explicitly. 
This is frequently denoted by \textit{active subspace learning}. Straightforward such approaches involve looking for linear trends using methods such as \textit{principal component analysis}~\cite{ulmasov2016bayesian} or \textit{partial least squares}~\cite{bouhlel2016improving}. \cite{djolonga2013high} use low-rank matrix recovery with directional derivatives with finite differences to find the active subspace, whereas~\cite{garnett2012embedding} learn a subspace using Bayesian active learning. If gradients are available, the active subspace is spanned by the eigenvectors of the matrix $C\coloneqq\int_{\mathcal{X}}\nabla f(x) (\nabla f(x))^Tdx$ with non-zero eigenvalues. This is used by \cite{constantine2015computing} and \cite{wycoff2021sequential} to show that $C$ can be estimated in closed form for \ac{GP} regression. We refer to the survey by~\cite{binois2022survey} for a more in-depth introduction to active subspace learning. Notably, large parts of the active subspace learning literature yield non-axis-aligned subspaces.

\subsection{Group testing}
Group testing~(GT,~\cite{aldridge2019group}) is a methodology for identifying elements with some low-probability characteristic of interest by jointly evaluating groups of elements.
\ac{GT} was initially developed to test for infectious diseases in larger populations but has later been applied in quality control~\cite{cuturi2020noisy}, molecular biology~\cite{balding1996comparative, ngo2000survey}, pattern matching~\cite{macula2004group, clifford2010pattern}, and machine learning~\cite{zhou2014parallel}. Group testing can be subdivided into two paradigms: \textit{adaptive} and \textit{non-adaptive}. Adaptive \ac{GT}, conducts tests sequentially, and previous results can influence the selection of subsequent groups. In the non-adaptive setting, the complete testing strategy is provided up-front. Furthermore a distinction can be made on whether test results are perturbed by evaluation noise. In the noisy setting, there is a risk that testing a group with active elements would show a negative outcome and vice versa. 
Our method presented in Section~\ref{sec:method} can be considered an adaptation of noisy adaptive \ac{GT}~\cite{scarlett2018noisy}.

\cite{cuturi2020noisy} present a \emph{Bayesian Sequential Experimental Design} approach for binary outcomes, which at each iteration selects groups that maximize one of two criteria: the first one is the mutual information between the elements' probability of being active, $\bm{\xi}$, in the selected group and the observation. 
The second is the area under the marginal encoder's curve~(AUC). 
As the distribution over the active group $p(\bm{\xi})$ is a $2^n$-dimensional vector, it quickly becomes impractical to store and update. Consequently, they propose using a \ac{SMC} sampler~\cite{del2006sequential}, representing the posterior probabilities by several weighted particles.

\section{Group testing for Gaussian Processes and Bayesian Optimization}\label{sec:method}
Our proposed method, \gtbo, leverages the assumption of axis-aligned active subspaces by explicitly identifying the active dimensions.
This gives the user additional insight into the problem and improves sample efficiency by focusing the optimization on the active dimensions.
This section describes how we adapt the \ac{GT} methodology and a BOED termination criterion to find active dimensions in as few evaluations as possible. Subsequently, we use the information to provide the surrogate model with the knowledge about which features are active.

\paragraph{Noisy adaptive group testing}
The underlying assumption is that a population of $n$ elements exists, each of which either possesses or lacks a specific characteristic. 
We refer to the subset of elements with this characteristic as the active group, considering the elements belonging to this group as active.
We let the \ac{RV} $\xi_i$ denote whether the element $i$ is active ($\xi_i=1$), or inactive ($\xi_i=0$), similar to \cite{cuturi2020noisy} who studied binary outcomes.
The state of the whole population can be written as the random vector $\bm{\xi} = \{\xi_1,\ldots,\xi_n\}\in\{0,1\}^n$.

We aim to uncover each element's activeness by performing repeated group tests.
We write $\bm{g}$ as a binary vector $\bm{g}=\{g_1,\ldots,g_n\}\in \{0,1\}^n$, such that $g_i=1$ signifies that element $i$ belongs to the group.
In noisy \ac{GT}, the outcome of testing a group is a random event described by the \ac{RV} $A(\bm{g}, \bm{\xi}) \in \{0,1\}$.
A common assumption is that the probability distribution of $A(\bm{g}, \bm{\xi})$ only depends on whether group $\bm{g}$ contains any active elements, i.e., $\bm{g}^\intercal\bm{\xi}\ge 1$.
In this case, one can define the sensitivity $p(A(\bm{g}, \bm{\xi}) =1~|~\bm{g}^\intercal\bm{\xi}\ge 1)$ and specificity $p(A(\bm{g}, \bm{\xi})=0~|~\bm{g}^\intercal\bm{\xi}= 0)$ of the test setup. As we assume the black-box function $f$ to be expensive to evaluate, we select groups $\bm{g}_t$ to learn as much as possible about the distribution $\bm{\xi}$ while limiting the number of iterations to $t=1\ldots T$, which subsequently limits the number of function evaluations.
We note that the group $\bm{g}_i$ is not a \ac{RV}, but is selected as part of \ac{GT} iteration $i$.

We can identify the active variables by modifying only a few variables in the search space and observing how the objective changes. 
Intuitively, if the function value remains almost constant after perturbing a subset of variables from the default point, it suggests that these variables are inactive. 
On the contrary, if a specific dimension $i$ is included in multiple subsets and the output changes significantly upon perturbation, this suggests that dimension $i$ is highly likely to be active.
 
Unlike in the traditional \ac{GT} problem, where outcomes are binary, we work with continuous, real-valued function observations.
To evaluate how a group of variables affects the objective function, we first evaluate a \emph{default} point in the center of the search space, $\bm{x}_\text{def}$, and then vary the variables in the group and study the difference.
We use the group notation $\bm{g}_t\in \{0,1\}^D$ as a binary indicator denoting which variables we change in iteration $t$. 
Similarly, we reuse the notation that the \ac{RV} $\bm{\xi}$ denotes the active dimensions, and the true state is denoted by $\bm{\xi}^*$. The new point to evaluate is selected as 
\begin{align}
    \bm{x}_t &= \bm{x}_{\rm def} \oplus (\bm{g}_t\otimes \bm{u}_t), \label{eq:point_creation}
\end{align}

where $\bm{u}_{t}\in [0,1]^D$ is drawn from $\mathcal{U}(\bm{0},\bm{1})$ until each active dimension has a distance of at least 0.4 to $\bm{x}_{\rm def}$, $\oplus$ is element-wise addition, and $\otimes$ is element-wise multiplication.
Note that a point $\bm{x}_t$ is always associated with a group $\bm{g}_t$ that determines along which dimensions $\bm{x}_t$ differs from the default point. For the newly obtained point $\bm{x}_t$, we must assess whether $|f(\bm{x_t})-f(\bm{x}_{\rm def})|\gg 0$, which would indicate that the group $\bm{g}_t$ contains active dimensions, i.e., $\bm{g}_t^\intercal \bm{\xi}^* \geq 1$. 
However, as we generally do not have access to the true values $f(\bm{x}_{\rm def})$ or $f(\bm{x}_{t})$ due to observation noise, we use an estimate $\hat{f}(\bm{x})$. Since $f$ can only be observed with Gaussian noise of unknown variance $\sigma_n^2$, there is always a non-zero probability that a high difference in function value occurs between $\bm{x}$ and $\bm{x}_\textrm{def}$ even if group $\bm{g}$ contains no active dimensions. Therefore, we take a probabilistic approach, which relies on two key assumptions:
\begin{enumerate}[leftmargin=*,noitemsep]
    \item $Z_t \coloneqq \hat{f}(\bm{x}_t) - \hat{f}(\bm{x}_{\rm def})\sim \mathcal{N}(0, \sigma_n^2)$ if $\bm{g}_t^\intercal \bm{\xi} = 0$, i.e., function values follow the noise distribution if the group $\bm{g}_t$ contains no active dimensions.
    \item $Z_t \coloneqq \hat{f}(\bm{x}_t) - \hat{f}(\bm{x}_{\rm def})\sim \mathcal{N}(0, \sigma^2)$ if $\bm{g}_t^\intercal\bm{\xi} \geq 1$, i.e., function values are drawn from a zero-mean Gaussian distribution with the function-value variance if the group $\bm{g}_t$ contains active dimensions.
\end{enumerate}

The first assumption follows from the assumption of Gaussian observation noise and an axis-aligned active subspace. 
The second assumption follows from a \ac{GP} prior assumption on $f$, under which $\hat{f}(x_t)$ is normally distributed.
As we are only interested in the change from $f(\bm{x}_\text{def})$, we assume this distribution to have mean zero. 

We estimate the noise variance, $\sigma_n^2$, and function-value variance, $\sigma^2$, based on an assumption on the maximum number of active variables.
First we evaluate $f$ at the default point $\bm{x}_{\text{def}}$. We then split the dimensions into several roughly equally sized bins.
For each bin, we evaluate $f$ on the default point perturbed along the direction of all variables in that bin and compare the result with the default value.
We then estimate the function variance as the empirical variance among the \texttt{max\_act} largest such differences and the noise variance as the empirical variance among the rest.
Here, \texttt{max\_act} represents the assumed maximum number of active dimensions.
If the assumption holds, there can be no active dimensions in the noise estimate, which is more sensitive to outliers.
It must be an upper bound, as the method is more sensitive to estimating the noise from active dimensions than vice versa.
An example of this is shown in Appendix~\ref{app:hyperparameters}.

Under Assumptions~1 and 2, the distribution of $Z_t$ depends only on whether $\bm{g}_t$ contains active variables.
Given the probability distribution over population states $p(\bm{\xi})$, the probability that $\bm{g}_t$ contains any active elements is
\begin{align}
    p(\bm{g}_t^\intercal \bm{\xi}\ge 1) = \sum_{\bm{\xi}\in \{0,1\}^D}\delta_{\bm{g}_t^\intercal \bm{\xi}\geq 1}p(\bm{\xi}). \label{eq:p_gamma_no_smc}
\end{align}

\begin{figure*}[bt!]
    \centering
    \includegraphics[width=.7\linewidth]{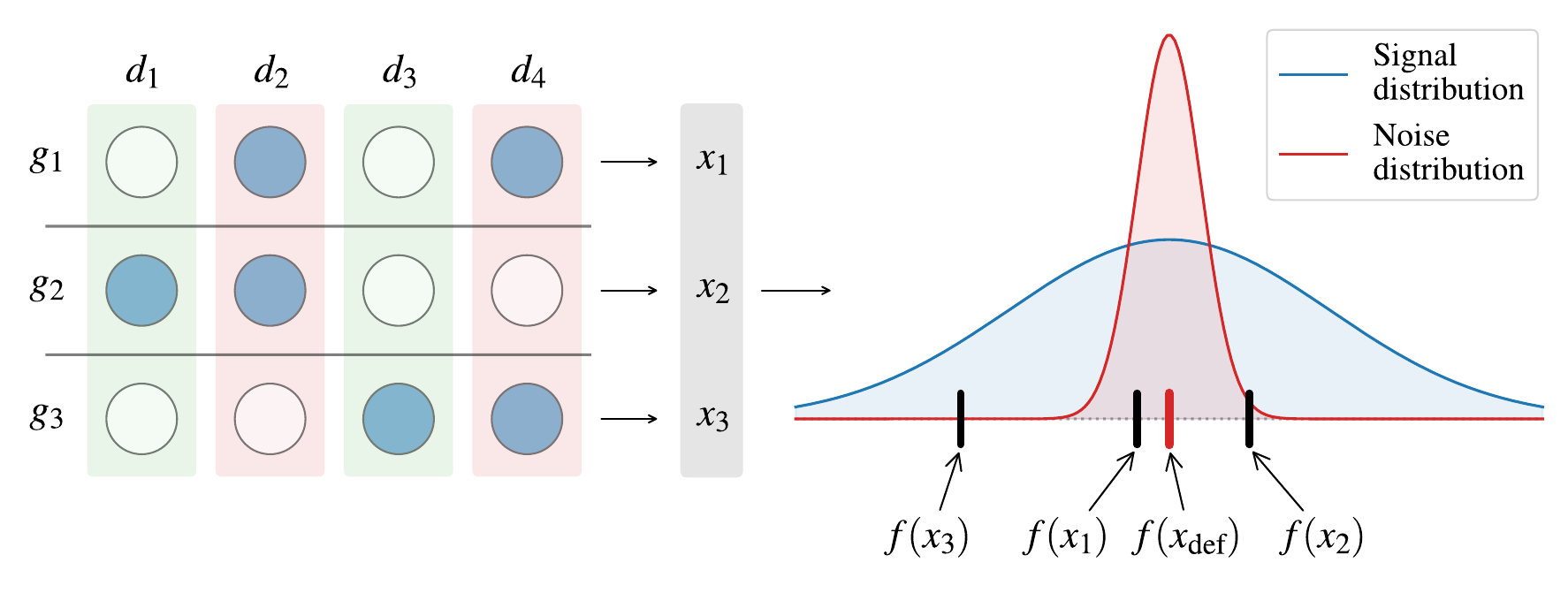}
    \caption{\gtbo assumes an axis-aligned subspace. A point $x_1$ that only varies along inactive dimensions ($d_2$ and $d_4$) obtains a similar function value as the default point $(x_\textrm{def})$. Points $x_2$ and $x_3$ that vary along active dimensions ($d_1$ and $d_3$) have a higher likelihood under the signal distribution than under the noise distribution.
    \vspace{-.35cm}
    }
    \label{fig:gtbo_scheme}
\end{figure*}

We exemplify this in Fig.~\ref{fig:gtbo_scheme}.
Here, three groups are tested sequentially, out of which the second and third contain active variables.
The three corresponding points, $x_1$, $x_2$, and $x_3$, give three function values shown on the right-hand side.
As observing $f(x_1)$ is more likely under the noise distribution, $g_1$ has a higher probability of being inactive.
Similarly, as $f(x_2)$ and $f(x_3)$ are more likely to be observed under the signal distribution, $g_2$ and $g_3$ are more likely to be active.

\paragraph{Estimating the group activeness probability}
Equation~\eqref{eq:p_gamma_no_smc} requires summing over $2^D$ possible activity states, which, for higher-dimensional functions, becomes prohibitively expensive.
Instead, we use an \ac{SMC} sampler with $M$ particles $\{\bm{\xi}_1, \ldots, \bm{\xi}_M\}$ and particle weights $\{\omega_1, \ldots, \omega_M\}$.
Each particle $\bm{\xi}_k \in \{0, 1\}^D$ represents a possible ground truth. 
We follow the approach presented in \cite{cuturi2020noisy} and use a modified Gibbs kernel for discrete spaces~\cite{liu1996peskun}. 
We then estimate the probability $p(\bm{g}_t^\intercal \bm{\xi}\ge 1)$ of a group $\bm{g}_t$ to be active by
\begin{align}
    \hat{p}(\bm{g}_t^\intercal \bm{\xi}\ge 1) = \sum_{k=1}^M \omega_k \delta_{\bm{g}_t^\intercal\bm{\xi}_k \geq 1}.
\end{align}

\paragraph{Choice of new groups}
We choose new groups to maximize the information obtained about $\bm{\xi}$ when observing $Z_t$.
This can be achieved by maximizing their \ac{MI}.
Under Assumptions~1 and 2, we can write the \ac{MI} as
\begin{align}
    I(\bm{\xi}, Z_t) &= H(Z_t)-H(Z_t|\bm{\xi}) \label{eq:mi} \\
    &= H(Z_t)- \sum_{\bm{\bar{\xi}}\in \{0,1\}^D} p(\bm{\bar{\xi}})H(Z_t|\bm{\xi}=\bm{\bar{\xi}})\\
    &= H(Z_t) -  [p(\bm{g}_t^\intercal \bm{\xi}\ge 1)H(Z_t | \bm{g}_t^\intercal \bm{\xi} \geq 1)\nonumber \\
    &\quad + p(\bm{g}_t^\intercal \bm{\xi}=0) H(Z_t | \bm{g}_t^\intercal \bm{\xi} =0) ] \\
    &= H(Z_t) - \frac{1}{2}[p(\bm{g}_t^\intercal \bm{\xi}=0)\log (2\sigma_n^2\pi e)\nonumber \\
    &\quad +p(\bm{g}_t^\intercal \bm{\xi}\ge 1)\log (2\sigma^2 \pi e)].
\end{align}

Since $Z_t$ is modeled as a \ac{GMM}, its entropy $H(Z_t)$ has no known closed-form expression~\cite{huber2008entropy}, but can be approximated using Monte Carlo:
\begin{align}
    H(Z_t) = \mathbb{E}[- \log p(Z_t)] \approx -\frac{1}{N}\sum_{i=1}^N\log p(z_t^i),
\end{align}
and $z_t^i \sim \mathcal{N}(0, \sigma^2)$ with probability $\hat{p}(\bm{g}_t^\intercal \bm{\xi}\ge 1)$ and $z_t^i \sim \mathcal{N}(0, \sigma_n^2)$ with probability $\hat{p}(\bm{g}_t^\intercal \bm{\xi} = 0)$.

\paragraph{Maximizing the mutual information}
\gtbo optimizes the \ac{MI} using a multi-start forward-backward algorithm~\cite{russell2010artificial}.
First, several initial groups are generated by sampling from the prior and the posterior over $\bm{\xi}$. 
Then, elements are greedily added for each group in a \textit{forward phase} and removed in a subsequent \textit{backward phase}. 
In the forward phase, we incrementally include the element that results in the greatest \ac{MI} increase.
Conversely, in the backward phase, we eliminate the element that contributes the most to \ac{MI} increase.
Each phase is continued until no further elements are added or removed from the group. 
Finally, the group with the largest \ac{MI} is returned.

\paragraph{Batch evaluations}
If the black-box function can be run in parallel, we greedily select additional groups by running the forward-backward algorithm again, excluding already selected groups. For high-dimensional problems there are frequently several distinct groups which each yields close to optimal \ac{MI}. We continue adding groups to evaluate until we have reached a user-specified upper limit or until the \ac{MI} of new groups drops below a threshold.

\paragraph{Updating the activeness probability}
Once we have selected a new group $\bm{g}_t$ and observed the corresponding function value $z_t$, we update our estimate of $\hat{p}(\bm{\xi}_k)$ for each particle $k$:
\begin{align}
    \hat{p}^{t}(\bm{\xi}_k)&\propto \hat{p}^{t-1}(\bm{\xi}_k)p(z_t|\bm{\xi}_k)\\
    &\propto \hat{p}^{t-1}(\bm{\xi}_k)
    \begin{cases} 
        p(z_t|\bm{g}_t^\intercal\bm{\xi}_k \geq 1) \,&\text{if}\,  \bm{g}_t^\intercal\bm{\xi}_k \geq 1 \\
        p(z_t|\bm{g}_t^\intercal\bm{\xi}_k = 0 ) \, &\text{if}\,  \bm{g}_t^\intercal\bm{\xi}_k = 0,
    \end{cases}
\end{align}
where $p(z_t|\bm{g}_t^\intercal\bm{\xi}_k = 0)$ and $p(z_t|\bm{g}_t^\intercal\bm{\xi}_k \geq 1)$ are Gaussian likelihoods.
Assuming that the probabilities of dimensions to be active are independent, the prior probability is given by ${\hat{p}^0(\bm{\xi}_k)= \prod_{i=1}^D q_i^{\bm{\xi}_{k,i}}(1-q_i)^{1-\bm{\xi}_{k,i}}}$ where $q_i$ is the prior probability for the $i$-th dimension to be active. 
As we represent the probability distribution $\hat{p}^0(\bm{\xi})$ by a point cloud, any prior distribution can be used to insert prior knowledge.
We use the same \ac{SMC} sampler as \cite{cuturi2020noisy}.

\paragraph{The GTBO algorithm}
With the individual parts defined, we present the complete procedure for \gtbo. 
\gtbo iteratively selects and evaluates groups for $T$ iterations or until convergence.
We consider it to have converged when the posterior marginal probability for each variable $\hat{p}^t(\xi_i)$ lies in $[0,C_\text{lower}]\cup [C_\text{upper},1]$, for some convergence thresholds $C_\text{lower}$ and $C_\text{upper}$.
More details on the \ac{GT} phase can be found in Algorithm~\ref{alg:full} in Appendix~\ref{app:algorithm}. Subsequently, their marginal posterior distribution decides which variables are selected to be active.
A variable $i$ is considered active if its marginal is larger than some threshold, $\hat{p}^t_i(\bm{\xi})\ge \eta$. 
Once we have deduced which variables are active, we perform \ac{BO} using the remaining sample budget.
To strongly focus on the active subspace, we use short lengthscale priors for the active variables and extremely long lengthscale priors for the inactive variables. Parameters lelated to the algorithm can be found in App~\ref{app:setup}.
We use a \ac{GP} with a Matérn$-\nicefrac{5}{2}$ kernel as the surrogate model and \texttt{qLogNoisyExpectedImprovement}~\cite{balandat2020botorch, ament2023unexpected} as the acquisition function.
The \ac{BO} phase is initialized with data sampled during the feature selection phase.
Several points are sampled throughout the \ac{GT} phase that only differ marginally in the active subspace.
Such duplicates are removed to facilitate the fitting of the \ac{GP}.

\section{Computational experiments}
In this section, we showcase the performance of the proposed methodology, both for finding the relevant dimensions and for the subsequent optimization.
We compare state-of-the-art frameworks for high-dimensional \ac{BO} on several synthetic and real-life benchmarks.
\gtbo outperforms previous approaches on the tested real-world and synthetic benchmarks. 
In Section~\ref{sec:sensitivity}, we study the sensitivity of \gtbo to external traits of the optimization problem, such as noise-to-signal ratio and the number of active dimensions.
The efficiency of the \ac{GT} phase is tested against other feature analysis algorithms in Appendix~\ref{app:comparison_feature_importance} and the \gtbo wallclock times are presented in Table~\ref{tab:runtimes} in Appendix~\ref{app:runtimes}.
We outline the full experimental setup in App.~\ref{app:setup}, and the code for \gtbo is available at \url{https://github.com/gtboauthors/gtbo}.

\subsection{Performance of the group testing}\label{ssec:performance_group_testing}
\label{sec:sensitivity}
\begin{figure}[t]
     \centering
    \includegraphics[width=0.85\linewidth]{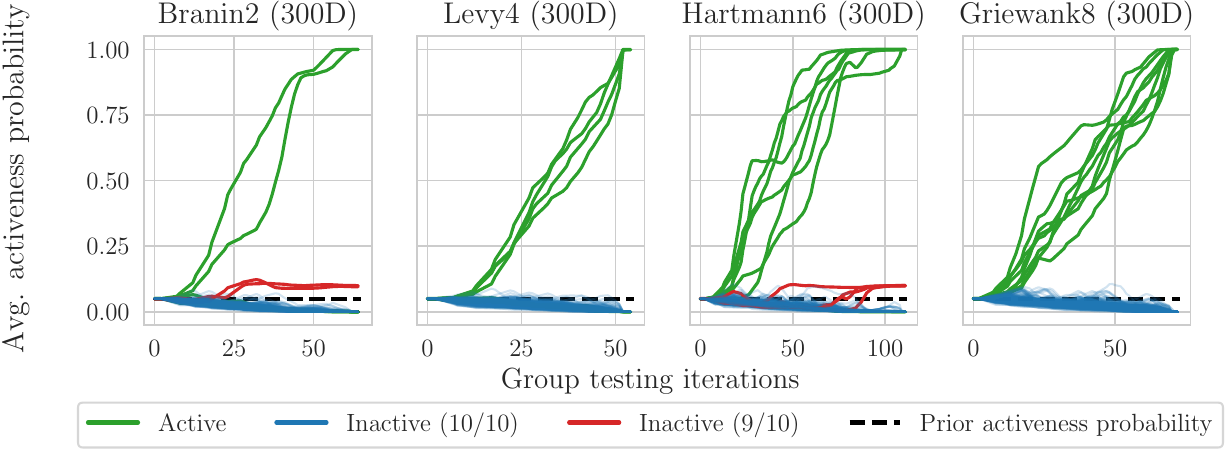}
    \caption{Evolution of the average marginal probability of being active across ten repetitions. Each line represents one dimension; active dimensions are colored green, and inactive dimensions are blue. In the few cases where \gtbo finds inactive variables to be active, the lines are emphasized in red. The last iteration marks the end of the \textit{longest} \ac{GT} phase across all runs. All active dimensions are identified in all runs. 6 out of 1180 inactive dimensions are incorrectly classified as active \textit{once} in ten runs across the benchmarks.
    \vspace{-.35cm}
    }
    \label{fig:synthetic_gt}
\end{figure}
Before studying \gtbo's overall optimization performance in high-dimensional settings, we analyze the performance of the \ac{GT} procedure.
In Fig.~\ref{fig:synthetic_gt}, we show the evolution of the average marginal probability of being active over the iterations for the different dimensions.
The truly active dimensions are plotted in green, and the inactive ones are in blue squares.
For all the problems, \gtbo correctly classifies all active dimensions during all runs within 39--112 iterations.
Across ten runs, \gtbo{} misclassifies 6 out of 1180 inactive variables to be active once each, for a false positive rate of 0.05\%.
The change in the number of active variables time is shown in Appendix~\ref{app:number_active}.

\paragraph{Sensitivity analysis}
\begin{figure}[t]
    \centering
    \includegraphics[width=0.85\linewidth]{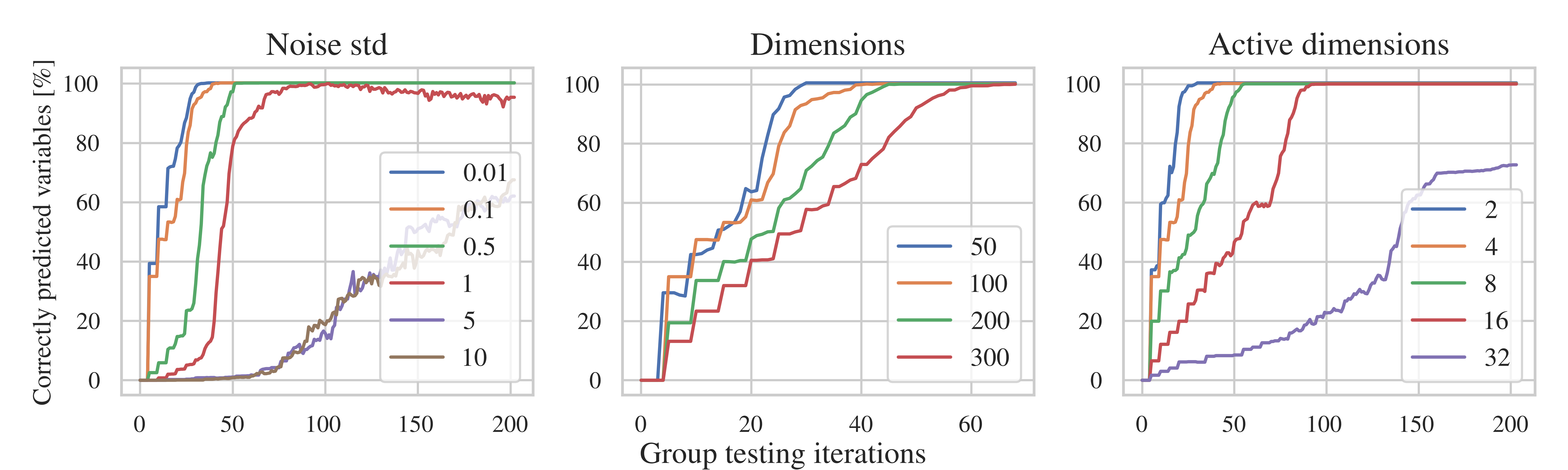}
     \caption{Sensitivity analysis for \gtbo. The average percentage of correctly classified variables is displayed for increasing \ac{GT} iterations. The percentage is ablated for (left) various levels of \textit{output noise}, (middle) number of \textit{total dimensions}, and (right) number of \textit{effective dimensions}. Each legend shows the points of the respective parameter.}
     \vspace{-.35cm}
    \label{fig:sens}
\end{figure}
We explore the sensitivity of \gtbo to the output noise and problem size by evaluating it on the \texttt{Levy4} synthetic benchmark extended to 100 dimensions, with a noise standard deviation of 0.1, and varying the properties of interest. 
In Fig.~\ref{fig:sens}, we show how the percentage of correctly predicted variables evolves with the number of tests $t$ for different functional properties. 
Correctly classified is defined here as having a probability of less than 1\% if inactive or above 90\% if active.
\gtbo shows to be robust to lower noise but suffers from very high noise levels.
As expected, higher function dimensionality and number of active dimensions increases the time until convergence.
Note that the signal and noise variance estimates build on the assumption that there are a maximum of $\sqrt{D}$ active dimensions, which does not hold with 32 active dimensions.
\begin{figure*}[t]
    \centering
    \includegraphics[width=.95\textwidth]{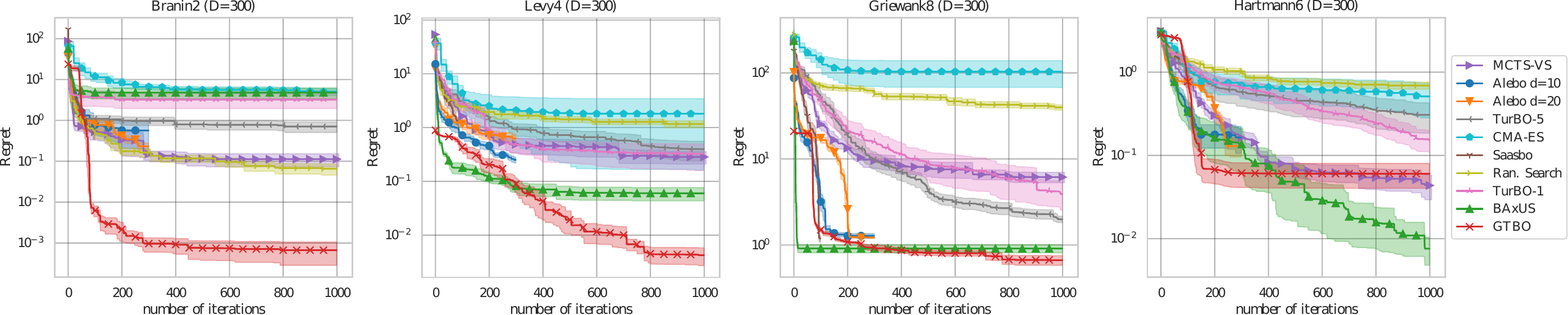}
     \caption{Mean logarithmic regret for four embedded synthetic benchmarks. The shaded regions indicate one standard error.
\gtbo finds active dimensions and subsequently optimizes efficiently.
     \vspace{-.35cm}
     }
    \label{fig:bsbo_runs_synth_71}
\end{figure*} 

\subsection{Optimization of real-world and synthetic benchmarks}\label{subsec:experiments}
\begin{figure}[t]
    \centering
    \includegraphics[width=0.8\linewidth]{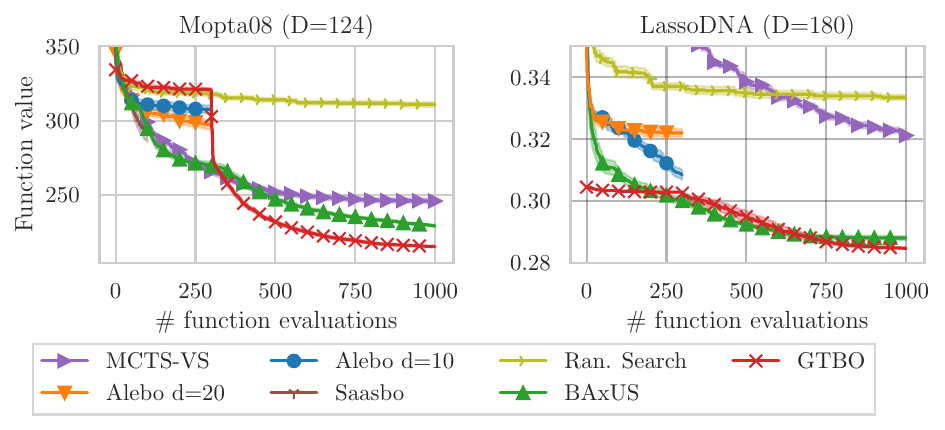}
     \caption{\gtbo outperforms competitors in real-world experiments. Notably, the performance on \texttt{Mopta08} increases significantly after the \ac{GT} phase at iteration 300, suggesting that the dimensions found during the \ac{GT} phase are highly relevant. 
     \vspace{-.35cm}
     }
    \label{fig:bsbo_runs_rw_31}
\end{figure}

We show that identifying the relevant variables can drastically improve optimization performance.
Fig.~\ref{fig:bsbo_runs_rw_31} shows the performance of \gtbo and competitors on the real-world benchmarks, Fig.~\ref{fig:bsbo_runs_synth_71} on the synthetic benchmarks.
The results show the incumbent function value for each method, averaged over ten repeated trials.
We plot the true average incumbent function values on the noisy benchmarks without observation noise. Since \texttt{Griewank} has its optimum in the center of the search space, we run \gtbo with a non-standard default away from the optimum.
However, the optimum being in the center means that all linear projections will contain the optimum, which boosts the projection-based methods \alebo and \baxus.

%
\new{Figure~\ref{fig:bsbo_runs_rw_31} shows \gtbo's performance on the 124D \texttt{Mopta08} and 180D \texttt{LassoDNA} benchmarks.
On \texttt{Mopta08}, \gtbo performs like a random search during the \ac{GT} phase but quickly outperforms the other methods once the \ac{BO} phase begins.
This behavior suggests that several dimensions of the \texttt{Mopta08} benchmark have negligible impact on the optimization objective and highlights \gtbo's ability to leverage the activeness information for benchmarks with inactive dimensions efficiently.

On \texttt{LassoDNA}, \gtbo again reaches similar performance levels as the random search and improves significantly when the \ac{BO} phase starts, outperforming \cmaes after $1\,000$ function evaluations. 
However, the overall performance is not on par with other \ac{BO} methods. 
Perhaps the fact that both variable selection methods, \mctvs and \gtbo, fail to reach the same performance levels as \turbo and \baxus suggests that most dimensions in \texttt{LassoDNA} are active. 
While this benchmark is suspected to violate our axis-aligned assumptions, \gtbo still shows reasonable performance and outperforms \mctvs by a large margin. Overall, \gtbo first identifies relevant variables, followed by a sharp drop when the optimization phase starts, indicating that knowing the active dimensions drastically speeds up optimization.
In the real-world \texttt{Mopta08} application where the dimensions detected as inactive have negligible impact on the objective, \gtbo outperforms state-of-the-art methods despite the delayed onset of the \ac{BO} phase. 
However, \gtbo can suffer in cases where the axis-alignment assumption does not hold, as shown by the \texttt{LassoDNA} results.}

\section{Discussion}
We propose \gtbo, a novel algorithm that focuses the \acl{BO} on variables found relevant in a preceding \acl{GT} phase. 
\gtbo explicitly exploits the structure of a sparse axis-aligned subspace to reduce the complexity of an application in high dimensions and is the first method to adapt \acl{GT}, in which it aims to find infected individuals by conducting pooled tests, to \acl{BO}.
It differs from \saasbo~\cite{eriksson2021high} in that it yields a clear-cut decision on which variables are active or inactive and from \alebo~\cite{letham2020re} or \baxus~\cite{papenmeier2022increasing} in that it does not rely on random projections to identify relevant variables.
Similarly to \mctvs~\cite{song2022monte}, our method explicitly identifies the set of relevant variables; however, \gtbo is the first method to use a more principled approach to learn them by employing Group Testing principles and theory.

\gtbo quickly detects active and inactive variables and shows robust optimization performance in synthetic and real-world settings.
Furthermore, the \ac{GT} phase yields a set of relevant dimensions, which allows users to learn something fundamental about their application. \gtbo robustly uses the activeness information so that it can still optimize efficiently even if the inactive dimensions have a marginal impact on the objective function. In future work, we intend to further investigate the impact of our \ac{GT} framework in the presence of noise.

\paragraph{Limitations} \gtbo has apparent shortcomings in its aggressive assumptions on the objective. Specifically, it relies on the assumption that an application has several irrelevant parameters, and that an existing subspace is axis-aligned. If these assumptions are is unmet, the method might underperform or waste a fraction of the evaluation budget to identify all variables as relevant.
\FloatBarrier
\newpage


\bibliography{bibliography/local,bibliography/lib,bibliography/proc,bibliography/strings}
\bibliographystyle{unsrt}

\appendix
\onecolumn
\section{The \gtbo Algorithm}
\label{app:algorithm}
This section describes the \ac{GT} phase of the \gtbo algorithm in additional detail.
\FloatBarrier
\begin{algorithm}[H]
\begin{algorithmic}
\REQUIRE black-box function~$f:\mathcal{X}\xrightarrow{}\mathbb{R}$, number of default point evaluations~$n_\text{def}$,  number of group tests~$T$, number of particles~$M$, prior distribution~$p^0(\bm{\xi})$
\ENSURE estimate of active dimensions $\bm{\gamma}$, posterior distribution $\hat{p}^T$
\FOR{$j\in \{1\ldots n_\text{def}\}$}
\STATE $y_\text{def}^{(i)} = y(\bm{x}_\text{def})$
\ENDFOR
\STATE $\hat{f}(\bm{x}_\text{def}) \gets \frac{1}{n_\text{def}}
\sum_{i=1}^{n_\text{def}} y_\text{def}^{(i)}$
\STATE
\STATE split the dimensions into $3\lfloor\sqrt{D}\rfloor$ bins $B$
\FOR{$j\in \{1\ldots|B|\}$}
\STATE $y_\text{bin}^{(j)} = y(\bm{x}_\text{def} \oplus \alpha b_j)$ \hfill $\triangleright \,\,$ a random perturbation along the dimensions in bin $b_j$
\ENDFOR
\STATE sort (ascending) $|y_\text{bin}^{(j)}-\hat{f}(\bm{x}_{\textrm{def}})|$
\STATE $\hat{\sigma}_n^2 \gets \texttt{var} (|y_\text{bin}^{(1)}-\hat{f}(\bm{x}_{\textrm{def}})|,\ldots,|y_\text{bin}^{(2\lfloor\sqrt{D}\rfloor)})-\hat{f}(\bm{x}_{\textrm{def}})|)$
\STATE $\hat{\sigma}^2 \gets \texttt{var} (|y_\text{bin}^{(2\lfloor\sqrt{D}\rfloor + 1)}-\hat{f}(\bm{x}_{\textrm{def}})|,\ldots,|y_\text{bin}^{(3\lfloor\sqrt{D}\rfloor)}-\hat{f}(\bm{x}_{\textrm{def}})|)$
\STATE
\STATE $\bm{\xi}_1, \ldots, \bm{\xi}_M \sim \hat{p}^0(\bm{\xi})$
\STATE $\omega_1, \ldots, \omega_M \gets \frac{1}{M}$ \hfill $\triangleright \,\,$ initial particle weights
\FORALL{$t \in \{1, \ldots, T\}$}
    \STATE $\bm{g}^* \gets $ \texttt{maximize\_mi}$( \bm{\xi}_1, \ldots, \bm{\xi}_M)$ \hfill $\triangleright \,\,$ find a group that maximizes \ac{MI}
    \STATE $\bm{x}_t \gets $ create using Eq.~\eqref{eq:point_creation} and $\bm{g}^*$
    \STATE $z_t \gets f(\bm{x}_t)+\epsilon - \hat{f}(\bm{x}_\text{def})$
    \STATE $(\bm{\xi}_i, \omega_i)_{i\in [M]}\gets \texttt{resample}(z_t,(\bm{\xi}_i, \omega_i)_{i\in [M]})$
    \STATE $\hat{p}^t \gets$ \texttt{marginal}$((\bm{\xi}_i, \omega_i)_{i\in [M]})$ \hfill $\triangleright \,\,$ get marginals
\ENDFOR
\STATE $\bm{\gamma} \gets (\delta_{\hat{p}^T_1(\bm{\xi}) \geq \eta}, \ldots, \delta_{\hat{p}^T_D(\bm{\xi}) \geq \eta})$ \hfill $\triangleright \,\,$ check which dimensions are active
\end{algorithmic}
\caption{Group testing phase}
\label{alg:full}
\end{algorithm}

\section{Comparison with feature importance algorithms}
\label{app:comparison_feature_importance}
We compare the performance of the \ac{GT} phase with the established feature importance analysis methods \xgboost~\cite{chen2016xgboost} and \fanova~\cite{hutter2014efficient}.
Since \fanova's stability degrades with increasing dimensionality, we run these methods on the 100-dimensional version of the synthetic benchmarks: \texttt{Branin2} (noise std 0.5), \texttt{Griewank8} (noise std 0.5), \texttt{Levy4} (noise std 0.1), and \texttt{Hartmann6} (noise std 0.01).

\begin{figure}[]
     \centering
     \begin{subfigure}[b]{0.48\textwidth}
         \centering
          \includegraphics[width=\textwidth]{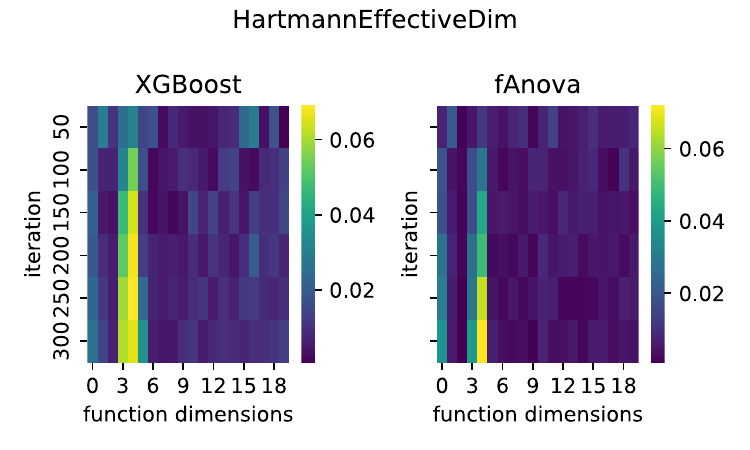}
          \caption{\texttt{Hartmann6}}
          \label{fig:feat_imp_hm6}
     \end{subfigure}
     \hfill
     \begin{subfigure}[b]{0.48\textwidth}
         \centering
          \includegraphics[width=\textwidth]{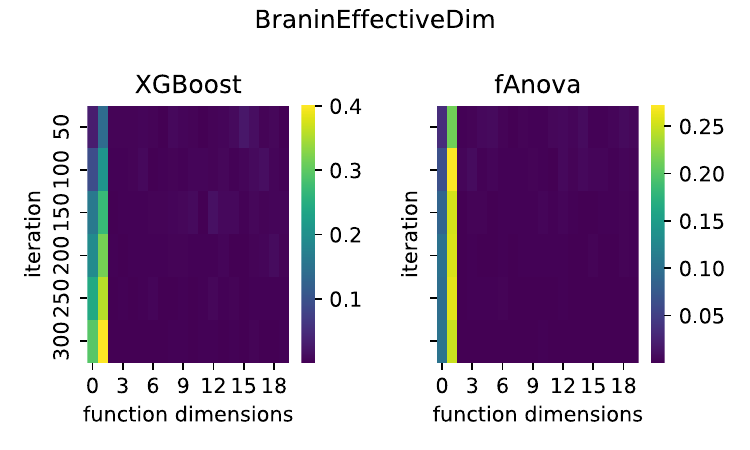}
          \caption{\texttt{Branin2}}
          \label{fig:feat_imp_brn2}
     \end{subfigure}
     \begin{subfigure}[b]{0.48\textwidth}
         \centering
          \includegraphics[width=\textwidth]{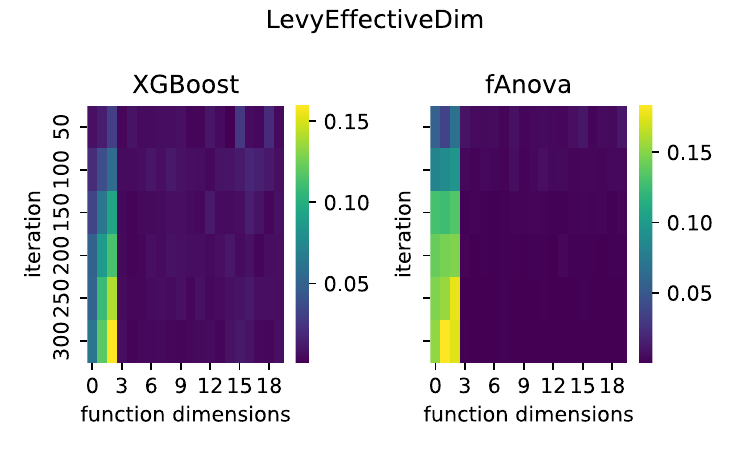}
          \caption{\texttt{Levy4}}
          \label{fig:feat_imp_levy4}
     \end{subfigure}
     \hfill
     \begin{subfigure}[b]{0.48\textwidth}
         \centering
          \includegraphics[width=\textwidth]{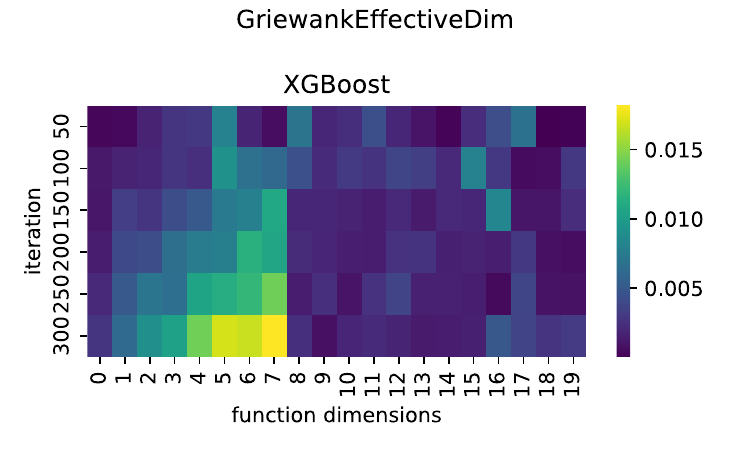}
          \caption{\texttt{Griewank8}}
          \label{fig:feat_imp_gw8}
     \end{subfigure}
     \caption{\xgboost and \fanova feature importance analysis on the 100-dimensional version of different synthetic benchmarks, averaged over 20 repetitions. Only the first 20 dimensions are shown. }
\end{figure}

Figure~\ref{fig:feat_imp_hm6} shows the results of \fanova and \xgboost on the 100-dimensional version of \texttt{Hartmann6} with added output noise. 
Per our results, both methods flag the third dimension as not more important than the added input dimensions (dimensions 7-100 with no impact on the function value).
Additionally, \fanova seems to ``switch off'' the second dimension.

On \texttt{Branin2} (Fig.~\ref{fig:feat_imp_brn2}), both methods detect the correct dimensions (the first and second dimensions).
Furthermore, all other dimensions have zero importance, and the methods find the correct partitioning earlier than for \texttt{Hartmann6}.
Similarly to \texttt{Hartmann6}, both methods fail to detect an active dimension (the fourth dimension).

\begin{figure}[]
    \centering
    \includegraphics[width=\textwidth]{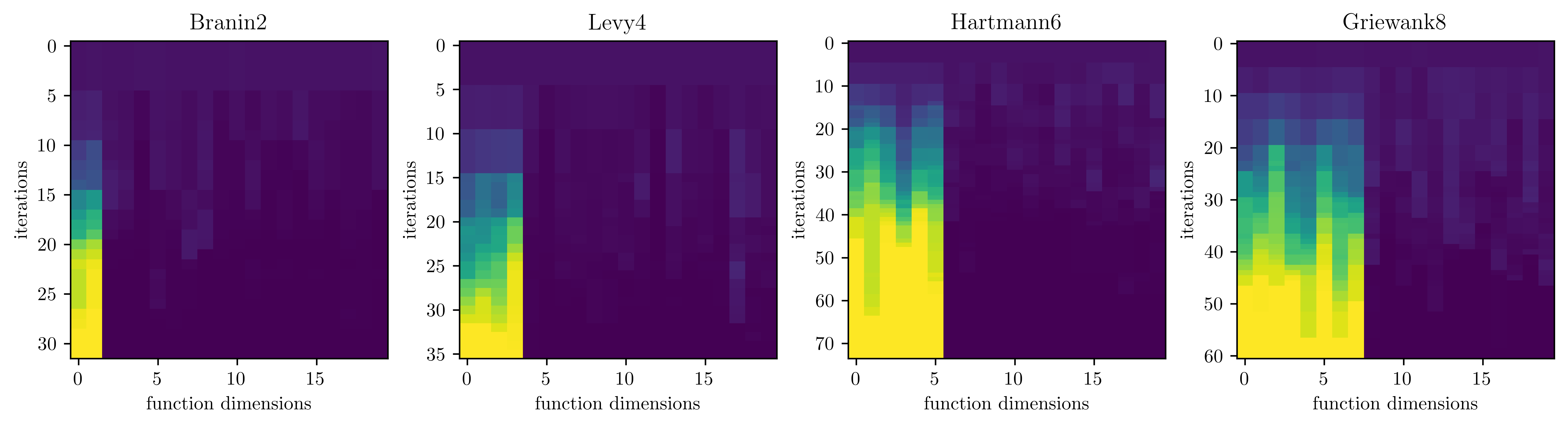}
    \caption{\gtbo-marginals of the first 20 dimensions., averaged over ten repetitions. If the \ac{GT} phase ends early, the last marginals are repeated to match the length of the longest \ac{GT} phase.}
    \label{fig:feat_gtbo_marg}
\end{figure}

On \texttt{Griewank8}, \fanova does not terminate gracefully.
Therefore, we only discuss \texttt{XGBoost} for \texttt{Griewank8}.
After 300 iterations, \texttt{XGBoost} only detects six dimensions reliably as active.
The other two dimensions are determined to be not more important than the added dimensions.
The marginals found by \gtbo are shown in Fig.~\ref{fig:feat_gtbo_marg}.
Compared to conventional feature importance analysis methods, \gtbo detects all active dimensions with high probability.

\section{Number of active variables}
\label{app:number_active}
Here, we show the average number of active dimensions throughout the \ac{GT} phase.
Given that the acceptance threshold of 0.5 is much higher than the initial probability of acceptance of 0.05, dimensions once considered active are rarely later considered inactive again, resulting in a close to monotonically increasing number of active dimensions.
\begin{figure}[]
    \centering
    \includegraphics[width=\textwidth]{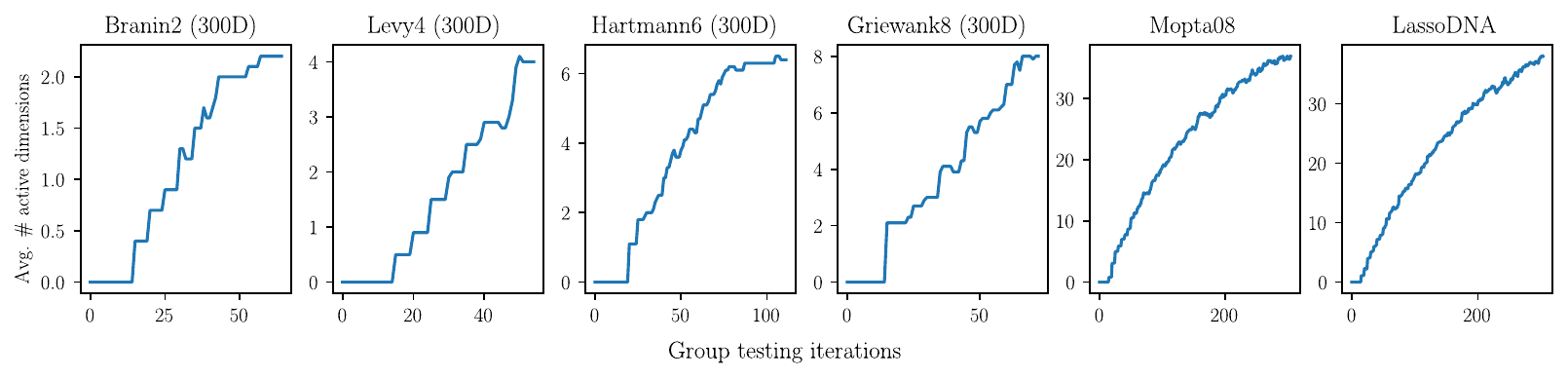}
    \caption{Evolution average number of active variables during the \ac{GT} phase (10 repetitions). The synthetic benchmarks find the correct number of active variables, whereas the real-world benchmarks find a significantly higher number.}
    \label{fig:avg_active}
\end{figure}

\section{Hyperparameter analysis}\label{app:hyperparameters}
\begin{figure}
    \centering
    \includegraphics[width = .8\textwidth]{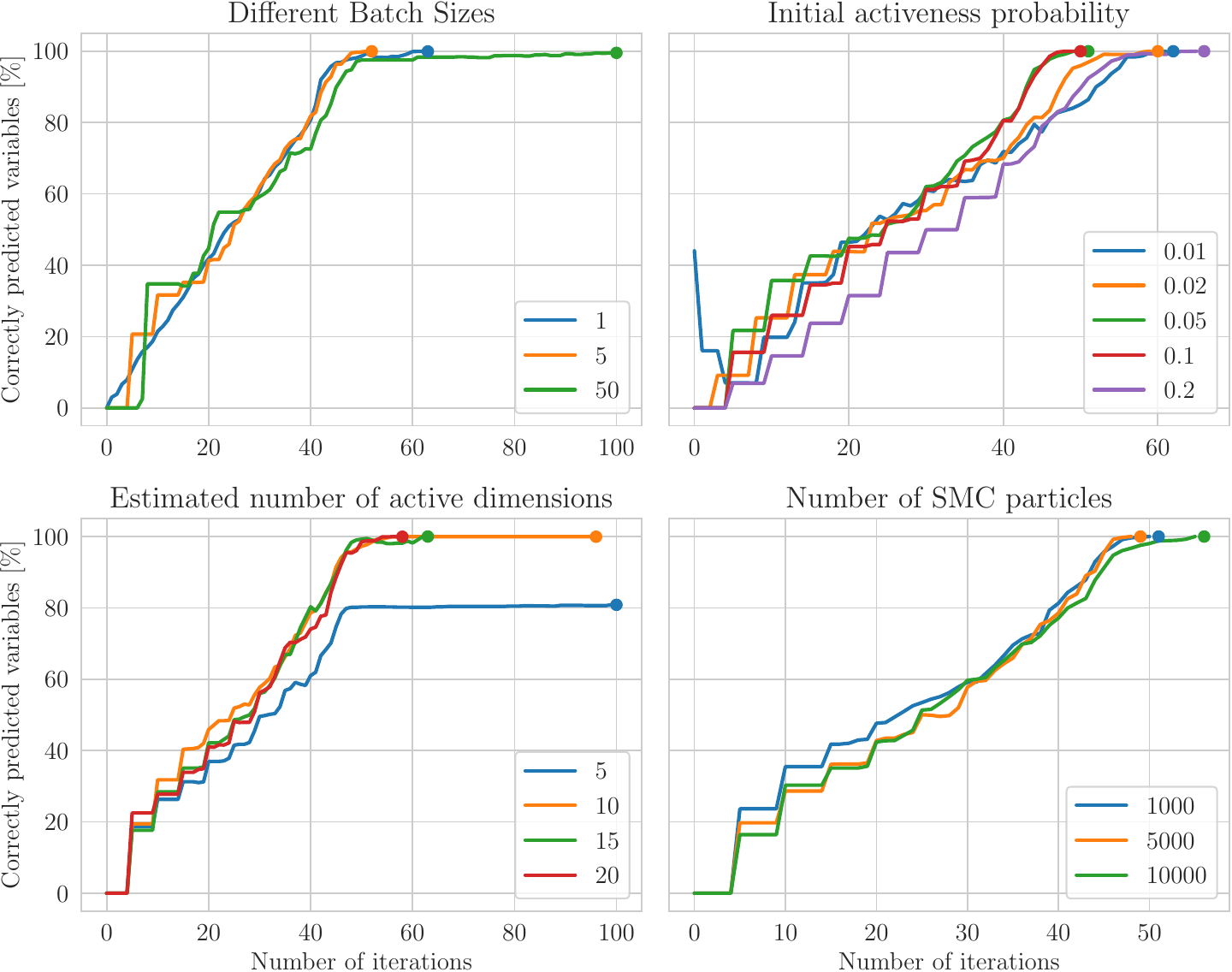}
    \caption{The percentage of correctly classified dimensions after an increasing number of iterations for different hyperparameter settings for Griewank8 (100D). The final dot on each line indicates that all runs have converged.}
    \label{fig:app_hp_sens}
\end{figure}

To measure the impact of different hyperparameters in \gtbo, we run an analysis on Griewank8 in 100 dimensions.
In Fig.~\ref{fig:app_hp_sens}, we focus on the maximum batch size, initial probability of being active, the assumed number of active dimensions for estimating the signal and noise ratio, and the number of particles in the \ac{SMC} sampler.
The iteration at which all 20 repetitions have converged for a single setting is marked with a circle.
Increasing the maximum batch size does not greatly impact the \ac{GT} performance.
The initial probability is important, but both 0.05 and 0.10 perform well; the very small or large values perform worse.
The assumed number of active dimensions, set to $\sqrt{D}$ in the paper, is highly important because it must be larger than the actual number of active dimensions.
Otherwise, active dimensions will be used to estimate the noise, severely hampering the performance.
We see this for the performance of the assumed active dimension of five, which is less than the true active dimensionality 8.
Lastly, the number of particles in the \ac{SMC} sampler is irrelevant for this benchmark, but it becomes more important for harder benchmarks.

\begin{figure}
    \centering
    \includegraphics[width = .5\textwidth]{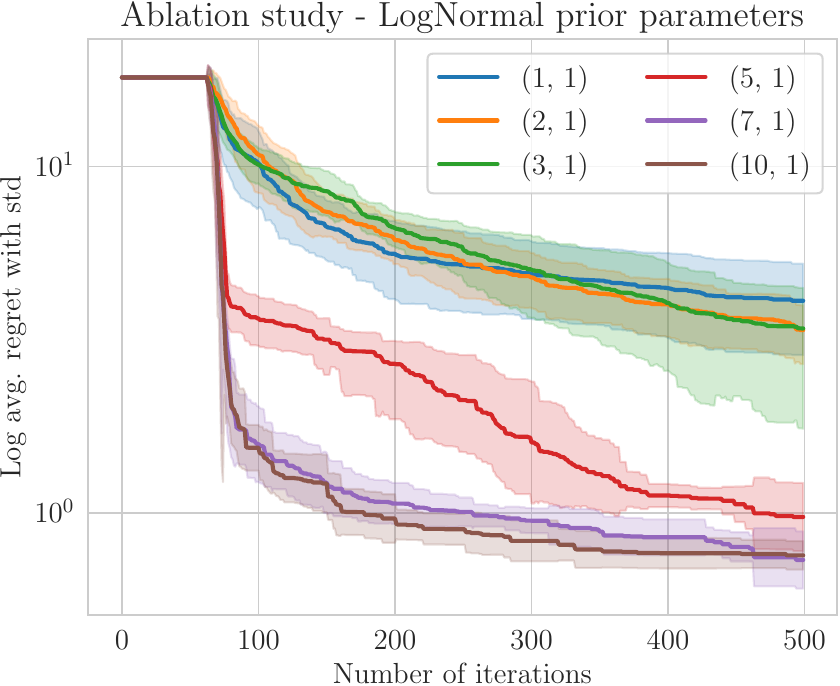}
    \caption{The log average regret of GTBO for different values of the LogNormal prior for Griewank8 (100D).}
    \label{fig:app_prior}
\end{figure}

We also study the impact of the parameters of the lengthscale LogNormal priors for inactive dimensions in Fig.~\ref{fig:app_prior}.
We see that the performance increases for longer lengthscales.
For the synthetic benchmarks, the inactive dimensions have absolutely no impact on the objective function, and as such, it is expected that de-emphasizing the importance of those variables would be beneficial.

\section{Experimental setup}\label{app:setup}

We test \gtbo on four synthetic benchmark functions, \texttt{Branin2}, \texttt{Levy} in 4 dimensions, \texttt{Hartmann6}, and \texttt{Griewank} in 8 dimensions, which we extend with inactive ``dummy'' dimensions~\cite{wang2016bayesian, eriksson2021high, papenmeier2022increasing} as well as two real-world benchmarks: the 124D soft-constraint version of the \texttt{Mopta08} benchmark~\cite{eriksson2021high}, and the 180D \texttt{LassoDNA} benchmark~\cite{vsehic2021lassobench}.
We add significant observation noise for the synthetic benchmarks, but the inactive dimensions are truly inactive.
In contrast, the real-world benchmarks do not exhibit observation noise, but all dimensions have at least a marginal impact on the objective function.  
Note that the noisy synthetic benchmarks are considerably more challenging for \gtbo than their noiseless counterparts.

Since the search space center is a decent solution for \texttt{LassoDNA}, \gtbo chooses a default point for each \ac{GT} repetition uniformly at random.
To not give \baxus a similar advantage, we subtract a random offset from the search space bounds, which we add again before evaluating the function.
This ensures that \baxus cannot always represent the near-optimal origin.

To evaluate the \ac{BO} performance, we benchmark against \turbo~\cite{eriksson2019scalable} with one and five trust regions, \saasbo~\cite{eriksson2021high}, \cmaes~\cite{hansen1996adapting}, \hesbo~\cite{nayebi2019framework}, and \baxus~\cite{papenmeier2022increasing} using the implementations and settings provided by the authors, unless stated otherwise.
We compare against random search, i.e., choose points in the search space uniformly at random.

We use the \texttt{pycma} implementation for \cmaes~\cite{nikolaus_hansen_2022_6370326} and the \texttt{Ax} implementation for \alebo~\cite{bakshy2018ae}.
To show the effect of different choices of the target dimensionality $d$, we run \alebo with $d = 10$ and $d = 20$. 
We observed that \alebo and \saasbo are constrained by their high runtime and memory consumption. 
The available hardware allowed up to 100 evaluations for \saasbo and 300 evaluations for \alebo for each run. 
Larger sampling budgets or higher target dimensions for \alebo resulted in out-of-memory errors. 
We note that limited scalability was expected for these two methods, whereas the other methods scaled to considerably larger budgets, as required for scalable \ac{BO}. We initialize each optimizer with ten initial samples and \baxus with $b = 3$ and $m_D = 1000$ and run ten repeated trials. 

Unless stated otherwise, we run \gtbo with $10\,000$ particles for the \ac{SMC} sampler, the prior probability of being active, $q=0.05$, and $3$ initial groups for the forward-backward algorithm. When estimating the function signal and noise variance, we set the assumed maximum number of active dimensions, \texttt{max\_act}, to $\sqrt{D}$. The threshold to be considered active after the \ac{GT} phase, $\eta$, is set to $0.5$, and the lower and upper convergence thresholds, $C_\textrm{lower}$ and $C_\textrm{upper}$, are $5\cdot 10^{-3}$ and $0.9$. We run all experiments with a log-normal $\mathcal{LN}(7,1)$ length scale prior to the inactive dimensions. If a benchmark is known to have strictly active and inactive parameters, this prior can be chosen more aggressively to ``switch off'' the inactive dimensions.
We use a $\mathcal{LN}(0,1)$ prior for the active variables, resulting in significantly shorter length scales.
In the \ac{GT} phase, we use batch evaluation with a maximum of 5 groups in each batch and a maximum \ac{MI} drop of 1\%.
Note that we still count the number of evaluations, not the number of batches, towards the budget. 
An analysis of the impact of some core hyperparameters is presented in Appendix~\ref{app:hyperparameters}.

The experiments are run on Intel Xeon Gold 6130 machines using two cores.

\section{Run times}\label{app:runtimes}
In this section, we show the average run of \gtbo.
Note that the \ac{SMC} resampling is a part of the \ac{GT} phase, and as such, the total algorithm time is the \ac{GT} time plus the \ac{BO} time.
We do batch evaluations in the \ac{GT} phase described in Section 3, with a maximum of five groups tested before resampling.
This significantly reduces the \ac{SMC} resample time.

\begin{table}[H]
    \centering
    \begin{tabular}{|l|lll|}
            \hline Benchmark & \ac{GT} time [h] & \ac{SMC} resample time [h] & \ac{BO} time [h] \\\hline
            \texttt{Branin2} (300D) & 1.94 & 0.583 & 9.31 \\
            \texttt{Levy4} (300D) & 2.33 & 0.603 & 10.8 \\
            \texttt{Hartmann6} (300D) & 3.29 & 1.14 & 14.1 \\
            \texttt{Griewank8} (300D) & 2.41 & 0.846 & 9.08 \\
            \texttt{Mopta08} (124D) & 5.70 & 4.74 & 7.95 \\
            \texttt{LassoDNA} (180D) & 8.92 & 7.06 & 10.2 \\\hline
    \end{tabular}
    \caption{Average \gtbo runtimes. Group testing time is on the same order of magnitude as the time allocated towards \ac{BO}. For \ac{BO}, the $\mathcal{O}(D^2)$ complexity of Quasi-Newton-based acquisition function optimization dominates the runtime.}
    \label{tab:runtimes}
\end{table}
\end{document}